\documentclass{article}
\usepackage{kr02}

\usepackage{times}
\usepackage{amsmath}

\title{Preferred well-founded semantics for logic programming\\ by alternating
  fixpoints: Preliminary Report}

\author{Torsten Schaub\thanks{~Affiliated with the School of Computing Science
  at Simon Fraser University, Burnaby, Canada.} \
  and 
  Kewen Wang\\
  Institut f\"ur Informatik,
  Universit\"at Potsdam\\
  Postfach 900327,
  D--14439 Potsdam,
  Germany\\
  \texttt{\{torsten,kewen\}@cs.uni-potsdam.de}
  }

\newcommand{\ClO}[0]{\ensuremath{\mathit{Cl}}}
\newcommand{\Cl}[1]{\ensuremath{\ClO(#1)}}

\newcommand{\lfp}{\ensuremath{\mathit{lfp}}} 
\newcommand{\gfp}{\ensuremath{\mathit{gfp}}} 

\newtheorem{definition}{Definition}
\newtheorem{theorem}{Theorem}
\newtheorem{corollary}[theorem]{Corollary}

\newcommand{\LPif}{\leftarrow}

\newcommand{\head}[1]{\ensuremath{\mathit{head}(#1)}}
\newcommand{\pbody}[1]{\ensuremath{\mathit{body}^+(#1)}}
\newcommand{\nbody}[1]{\ensuremath{\mathit{body}^-(#1)}}
\newcommand{\body}[1]{\ensuremath{\mathit{body}(#1)}}

\newcommand{\nafo}[0]{\mathit{not}}
\newcommand{\naf}[0]{\nafo\;}

\newcommand{\reduct}[2]{\ensuremath{#1^{#2}}}
\newcommand{\reductr}[1]{\ensuremath{#1^{+}}}

\newcommand{\To}[1]{\ensuremath{T_{#1}}}
\newcommand{\T}[2]{\To{#1}#2}
\newcommand{\TiO}[2]{\To{#2}^{#1}}
\newcommand{\Ti}[3]{\TiO{#1}{#2}#3}

\newcommand{\TPo}[3]{\ensuremath{\mathcal{T}_{(#1,#2),#3}}}
\newcommand{\TP}[4]{\TPo{#1}{#2}{#3}#4}
\newcommand{\TPX}[4]{\TP{#1}{#2}{#3}{#4}}

\newcommand{\TPiO}[4]{\TPo{#2}{#3}{#4}^{#1}}
\newcommand{\TPi}[5]{\TPiO{#1}{#2}{#3}{#4}#5}

\newcommand{\CnO}[0]{\ensuremath{\mathit{Cn}}}
\newcommand{\Cn}[1]{\ensuremath{\CnO(#1)}}
\newcommand{\CPO}[2]{\ensuremath{\mathcal{C}_{(#1,#2)}}}
\newcommand{\CPX}[3]{\ensuremath{\CPO{#1}{#2}(#3)}}
\newcommand{\CPOdefault}{\CPO{\Pi}{<}}
\newcommand{\CPXdefault}{\CPX{\Pi}{<}{X}}

\newcommand{\APO}[2]{\ensuremath{\mathcal{A}_{(#1,#2)}}}
\newcommand{\APX}[3]{\ensuremath{\APO{#1}{#2}(#3)}}
\newcommand{\APOdefault}{\APO{\Pi}{<}}
\newcommand{\APXdefault}{\APX{\Pi}{<}{X}}

\newcommand{\TPNo}[3]{\ensuremath{\mathcal{T}^\circ}_{(#1,#2),#3}}
\newcommand{\TPN}[4]{\TPNo{#1}{#2}{#3}#4}

\newcommand{\TPNiO}[4]{(\TPNo{#2}{#3}{#4})^{#1}}
\newcommand{\TPNi}[5]{\TPNiO{#1}{#2}{#3}{#4}#5}
\newcommand{\CPNO}[2]{\ensuremath{\mathcal{C}^\circ_{(#1,#2)}}}
\newcommand{\CPNX}[3]{\ensuremath{\CPNO{#1}{#2}(#3)}}
\newcommand{\CPNOdefault}{\CPNO{\Pi}{<}}
\newcommand{\CPNXdefault}{\CPNX{\Pi}{<}{X}}
\newcommand{\APNO}[2]{\ensuremath{\mathcal{A}^\circ_{(#1,#2)}}}

\newcommand{\APNOdefault}{\APNO{\Pi}{<}}

\newcommand{\Lit}[0]{\ensuremath{\mathit{Lit}}}

\begin{document}

\maketitle

\begin{abstract}
We analyze the problem of defining well-founded semantics for ordered logic
programs within a general framework based on alternating fixpoint theory.
We start by showing that generalizations of existing answer set approaches to
preference are too weak in the setting of well-founded semantics.
We then specify some informal yet intuitive criteria and propose a semantical
framework for preference handling that is more suitable for defining
well-founded semantics for ordered logic programs.
The suitability of the new approach is convinced by the fact that many
attractive properties are satisfied by our semantics.
In particular, our semantics is still correct with respect to various existing
answer sets semantics while it successfully overcomes the weakness of their
generalization to well-founded semantics.
Finally, we indicate how an existing preferred well-founded semantics
can be captured within our semantical framework.
\end{abstract}


\paragraph{Keywords:}
well-founded semantics, 
preference, 
alternating fixpoints,
extended logic programs.

\section{Introduction}
\label{sec:introduction}

Preferences constitute a very natural and effective way of resolving
indeterminate situations.
For example, in scheduling not all deadlines may be simultaneously
satisfiable,
and in configuration various goals may not be simultaneously met.
Preferences among deadlines and goals may allow for an acceptable,
non-optimal solution.
In legal reasoning, laws may apply in different situations, but laws may also
conflict with each other.
Conflicts are resolved by appeal to higher-level principles such as authority
or recency.
So federal laws will have a higher priority than state laws, and newer laws
will take priority over old.
Further preferences, such as authority holding sway over recency, may also
be required.
In fact,
while logical preference handling constitutes already an indispensable means
in legal reasoning systems (cf.~\cite{gordon93a,prakken97a}),
it is also advancing in other application areas
such as
intelligent agents and e-commerce \cite{grosof99a},
information-site selection~\cite{eifisato01a},
and
the resolution of grammatical ambiguities~\cite{cuiswi01}.

The increasing practical interest in preferences is also reflected by the
large number of proposals for preference handling in logic programming,
including
\cite{sakino96,brewka96a,gelson97a,zhafoo97a,grosof97,breeit99a,descto00c,wazhli00},
and related areas, such as default logic~\cite{baahol93a,brewka94a,delsch97a}.
A common approach in such work has been to employ meta-formalisms for
characterizing ``preferred answer sets''.
This has led to a diversity of approaches that are hardly comparable
due to considerably different methods of formal characterization.
As a consequence, there is no homogeneous account of preference.

In \cite{schwan01b},
we started addressing this shortcoming by proposing a uniform semantical
framework for extended logic programming with preferences.
To be precise,
we develop an (alternating) fixpoint theory for so-called
\emph{ordered logic programs},
building on the basic ideas in~\cite{vangelder93}.
An {ordered} logic program is an extended logic program whose rules are
subject to a strict partial order.
In analogy to standard logic programming, such a program is then interpreted
by means of an associated fixpoint operator.
Different semantics are obtained by distinguishing different subsets of the
respective set of alternating fixpoints.
As a result, several different approaches to defining preferred answer sets,
including~\cite{breeit98a,bufale99a,descto00c}, can all be captured within
our framework and each of these preference strategies is based on an
operator, which plays the same role as the consequence operator
in the setting of normal logic programs.

In this paper, we show that the preference strategies for defining answer sets
turn out to be too weak in the setting of well-founded semantics.
For this reason, we propose a new approach to preference handling for logic
programs that seems to be more appropriate for well-founded semantics. 
In fact, we show that for a resulting instance of this approach some attractive properties.  
We also discuss the relation of our preferred well-founded semantics to other
approaches~\cite{nute87,brewka96a,yowayu01}.


\section{Definitions and notation}
\label{sec:background}

%
An \emph{extended logic program} is a finite set of rules of the form
\begin{equation}\label{eqn:rule}
L_0\LPif L_1,\dots,L_m,\naf L_{m+1},\dots,\naf L_n,
\end{equation}
where $n\geq m\geq 0$, and each $L_i$ $(0\leq i\leq n)$ is a \emph{literal},
ie.\ either an atom $A$ or the negation $\neg A$ of $A$.
The set of all literals is denoted by $\Lit$.
Given a rule $r$ as in~(\ref{eqn:rule}),
we let $\head{r}$ denote the \emph{head}, $L_0$, of $r$
and
$\body{r}$ the \emph{body},
\(
\{L_1,\dots,L_m,\ \naf L_{m+1},\dots,\naf L_n\}
\),
of $r$.
Further, let
\(
\pbody{r}
=
\{L_1,,\dots, L_m\}
\)
and
\(
\nbody{r}
=
\{L_{m+1},\dots, L_n\}
\).
A program is called \emph{basic} if $\nbody{r}=\emptyset$ for all its rules;
it is called \emph{normal} if it contains no classical negation symbol $\neg$.

We define the reduct of a rule $r$ as $\reductr{r}=\head{r}\LPif\pbody{r}$.
The \emph{reduct}, $\reduct{\Pi}{X}$, of a program $\Pi$ \emph{relative to} a
set $X$ of literals is defined by
\[
\reduct{\Pi}{X}
=
\{\reductr{r}\mid r\in\Pi\text{ and }\nbody{r}\cap X=\emptyset\}.
\]
A set of literals $X$ is \emph{closed under} a basic program $\Pi$ iff for any
$r\in\Pi$, $\head{r}\in X$ whenever $\pbody{r}\subseteq X$.
We say that $X$ is \emph{logically closed} iff it is either consistent 
(ie.\ it does not contain both a literal $A$ and its negation $\neg A$) or
equals $\Lit$.
The smallest set of literals which is both logically closed and closed under a
basic program $\Pi$ is denoted by $\Cn{\Pi}$.
With these formalities at hand,
we can define \emph{answer set semantics} for extended logic programs:
A set $X$ of literals is an \emph{answer set} of a program $\Pi$
iff
$\Cn{\reduct{\Pi}{X}}=X$.

For capturing even more semantics within a similar framework,
van~Gelder defines in~\cite{vangelder93}
the operator $C_{\Pi}(X)$ as $\Cn{\Pi^X}$.
It is important to note that the operator $C_{\Pi}$ is anti-monotonic,
which implies that the operator $A_\Pi(X)=C_{\Pi}(C_{\Pi}(X))$
is monotonic. 
A fixpoint of $A_\Pi$ is called an \emph{alternating fixpoint} for $\Pi$.
Different semantics are captured by distinguishing different groups of fixpoints of
$A_\Pi$.
For instance, given a program $\Pi$,
the least alternating fixed point of $A_\Pi$ is known to amount to its
\emph{well-founded semantics}.
Answer sets of $\Pi$ are simply alternating fixed points of $A_\Pi$ that are
also fixed points of $C_{\Pi}$.

Alternative inductive characterizations for
the operators \CnO, $C_\Pi$, and $A_\Pi$ can be obtained by appeal to
\emph{immediate consequence operators}~\cite{vankow76,lloyd87}.
Let $\Pi$ be a basic program and $X$ a set of literals.
The \emph{immediate consequence operator} $\To{\Pi}$ is defined as follows:
\[
\T{\Pi}{X} = \{\head{r}\mid r\in\Pi\text{ and }\body{r}\subseteq X\}
\]
if $X$ is consistent, and $\T{\Pi}{X} = \Lit$ otherwise.
Iterated applications of $\To{\Pi}$ are written as $\TiO{j}{\Pi}$ for
$j\geq 0$, where
\(
\Ti{0}{\Pi}{X}=X
\)
and
\(
\Ti{i}{\Pi}{X}=\T{\Pi}{\Ti{i-1}{\Pi}{X}}
\)
for $i\geq 1$.
It is well-known that
\(
\Cn{\Pi}=\bigcup_{i\geq 0}\Ti{i}{\Pi}{\emptyset}
\),
for any basic program $\Pi$.
Also, for any answer set $X$ of program $\Pi$, it holds that
\(
X=\bigcup_{i\geq 0}\Ti{i}{\reduct{\Pi}{X}}{\emptyset}
\).

A reduction from extended to basic programs is avoidable with an extended
consequence operator:
Let $\Pi$ be an extended program and $X$ and $Y$ be sets of literals.
The \emph{extended immediate consequence operator} \To{\Pi,Y} is defined as
follows:
\begin{eqnarray}
  \label{eq:def:extended:immediate:consequence:operator}
  \T{\Pi,Y}{X}
  &=&
  \{\head{r}\mid r\in\Pi
                 , \ 
                   \pbody{r}\subseteq X, \\
  \nonumber     && \qquad\qquad\qquad\text{ and } 
                   \nbody{r}\cap Y=\emptyset
  \}
\end{eqnarray}
if $X$ is consistent, and $\T{\Pi,Y}{X} = \Lit$ otherwise.
Iterated applications of $\To{\Pi,Y}$ are written as those of $\To{\Pi}$ above.
Clearly, we have
\(
\T{\Pi,\emptyset}{X}=\T{\Pi}{X}
\)
for any basic program $\Pi$
and
\(
\T{\Pi,Y}{X}=\T{\Pi^Y}{X}
\)
for any extended program $\Pi$.
Accordingly, we have for any answer set $X$ of program $\Pi$ that
\(
X=\bigcup_{i\geq 0}\Ti{i}{\Pi,X}{\emptyset}
\).
Finally,
for dealing with the individual rules
in~(\ref{eq:def:extended:immediate:consequence:operator}),
we rely on the notion of \emph{activeness}:%
\footnote{Although \emph{activeness} is implicitly present in standard logic
  programming (cf.~definition of $\T{\Pi,Y}{X}$),
  the term as such was (to the best of our knowledge) coined in approaches
  dealing with preferences in default logic \cite{baahol93a,brewka94a}.
  There, however, activeness additionally stipulated that $\head{r}\not\in X$
  in order to prevent multiple applications of the same rule.}
Let $X,Y\subseteq\Lit$ be two sets of literals in a program $\Pi$.
A rule $r$ in $\Pi$ is \emph{active} wrt the pair $(X,Y)$,
if $\pbody{r}\subseteq X$ and $\nbody{r}\cap Y=\emptyset$.
Alternatively, we thus have that
\(
\T{\Pi,Y}{X}
=
\{\head{r}\mid r\in\Pi\text{ is active wrt }(X,Y)\}
\).

Lastly,
an \emph{ordered logic program}\footnote{Also called
  \emph{prioritized} logic program by some authors, as eg.\ in
  \cite{zhafoo97a,breeit99a}.}
is simply a pair $(\Pi,<)$, where $\Pi$ is an extended logic program and
$<\;\subseteq{\Pi\times\Pi}$ is an irreflexive and transitive relation.
Given, $r_1,r_2\in\Pi$, the relation $r_1<r_2$ is meant to express that $r_2$
has \emph{higher priority} than $r_1$.\footnote{Some authors,
  eg.\ \cite{breeit99a}, 
  attribute relation $<$ the inverse meaning.}


\section{Preferred (alternating) fixpoints}\label{sec:afp}

We start by describing the semantical framework given in \cite{schwan01b},
while concentrating on the formal details needed for capturing the approach
introduced in~\cite{wazhli00}.
The formal development of the approach in \cite{breeit99a} and
\cite{descto00c} is analogous and thus omitted here.

The overall idea behind the obtained semantics for ordered logic program is to
distinguish the ``preferred'' answers of a program $(\Pi,<)$ by means of
fixpoint equations.
That is, a set of literals $X$ constitutes a collection of preferred answers
from $(\Pi,<)$,
if it satisfies the equation
\(
\CPXdefault=X
\)
for some operator \CPOdefault.
In view of the classical logic programming approach described in
Section~\ref{sec:background},
this makes us investigate semantics that interpret preferences as inducing
selection functions on the set of standard answer sets of the underlying
non-ordered program $\Pi$.

Standard answer sets are defined via a reduction of extended logic programs
to basic programs.
Such a reduction is inappropriate when resolving conflicts among rules by
means of preferences since all such conflicts are simultaneously resolved when
turning $\Pi$ into $\reduct{\Pi}{X}$.
Rather conflict resolution must be addressed among the original rules in
order to account for blockage between rules.
In fact, once the negative body \nbody{r} is eliminated there is no way to
detect whether $\head{r'}\in\nbody{r}$ holds in case of $r<r'$.
Our idea is therefore to characterize preferred answer sets by an inductive
development that agrees with the given ordering rather than a simultaneous
reduction.
In terms of a standard answer set $X$, this means that we favor its formal
characterization as
\(
X=\bigcup_{i\geq 0}\Ti{i}{\Pi,X}{\emptyset}
\) 
over $X=\Cn{\reduct{\Pi}{X}}$.
This leads us to the following definition.  
\footnote{Fixpoint operators for the approaches in \cite{breeit99a} and
  \cite{descto00c} are obtained by appropriate modifications to Condition~I
  and~II in Definition~\ref{def:Tp:W}; cf.~\cite{schwan01b}.}
%
\begin{definition}\label{def:Tp:W}
Let $(\Pi,<)$ be an ordered logic program and let $X$ and $Y$ be sets of literals.

We define the set of 
immediate consequences of $X$ with respect to $(\Pi,<)$ and $Y$
as
\[
\TP{\Pi}{<}{Y}{X}
=
\left\{
  \head{r}\left|
    \begin{array}{cl}
      \mathit{I}. & r 
                    \text{ active wrt } (X,Y); 
      \\
      \mathit{II}. & \text{there is no rule }r' \\ 
                   & \text{ with } r< r'
                   \text{such that}
                     \\
                   & (a)\ r'
                        \text{ active wrt } (Y,X) 
                        \\
                   & (b)\ \head{r'}\not\in X
    \end{array}
  \right\}\right.
\]

if $X$ is consistent, and $\TPX{\Pi}{<}{Y}{X}=\Lit$ otherwise.
\end{definition}
%
Note that \TPo{\Pi}{<}{Y} is a refinement of its classical counterpart
\To{\Pi,Y}.
To see this, observe that Condition~\emph{I} embodies the standard
application condition for rules
given in~(\ref{eq:def:extended:immediate:consequence:operator})

The actual refinement takes place in Condition~\emph{II}.
The idea is to apply a rule $r$ only if the ``question of applicability'' has been
settled for all higher-ranked rules $r'$.
Let us illustrate this in terms of iterated applications of $\TPo{\Pi}{<}{Y}$.
In these cases, $X$ contains the set of conclusions that have been derived so
far, while $Y$ provides the putative answer set
(or: $\Lit\setminus Y$ provides a set of literals that can be falsified).
Then, the ``question of applicability'' is considered to be settled for a higher
ranked rule $r'$
\begin{itemize}
\item if the prerequisites of $r'$ will never be derivable,
viz.\ $\pbody{r'}\not\subseteq Y$,
\quad or
\item if $r'$ is defeated by what has been derived so far,
viz.\ $\nbody{r}\cap X\neq\emptyset$,
\quad or 
\item if $r'$ or another rule with the same head have already applied,
viz.\ $\head{r'}\in X$.
\end{itemize}
The first two conditions show why activeness of $r'$ is stipulated wrt
$(Y,X)$, as opposed to $(X,Y)$ in Condition~\textit{I}.
The last condition serves somehow two purposes:
First, it detects whether the higher ranked rule $r'$ has applied and,
second, it suspends the preference $r<r'$ whenever the head of the higher
ranked has already been derived by another rule.
This suspension of preference constitutes a distinguishing feature of the
approach at hand;
this is discussed in detail in \cite{schwan01b} in connection with
other approaches to preference handling.

As with $\To{\Pi}$ and $\To{\Pi,Y}$,
iterated applications of $\TPo{\Pi}{<}{Y}$ are written as
$\TPiO{j}{\Pi}{<}{Y}$ for $j\geq 0$, where
\(
\TPi{0}{\Pi}{<}{Y}{X}=X
\)
and
\(
\TPi{i}{\Pi}{<}{Y}{X}=\TP{\Pi}{<}{Y}{\TPi{i-1}{\Pi}{<}{Y}{X}}
\)
for $i\geq 1$.
This allows us to define the counterpart of fixpoint operator $C_\Pi$ for
ordered programs:
%
\begin{definition}\label{def:fixpoint:operator:W}
Let $(\Pi,<)$ be an ordered logic program and let $X$ be a set of literals.

We define
\(
\CPXdefault=\bigcup_{i\geq 0}\TPi{i}{\Pi}{<}{X}{\emptyset}
\).
\end{definition}
%
In analogy to \TPo{\Pi}{<}{Y} and \To{\Pi,Y},
operator \CPO{\Pi}{<} is a refinement of its classical counterpart $C_\Pi$.
The major difference of our definition from van~Gelder's is that we directly
obtain the consequences from  $\Pi$ (and $Y$).
Unlike this, the usual approach (without preferences) first obtains a basic
program \reduct{\Pi}{Y} from $\Pi$ and then the consequences are derived from
this basic program \reduct{\Pi}{Y}.

A preferred answer set is defined as a fixpoint of \CPOdefault.

In analogy to van~Gelder~\cite{vangelder93}, we may define the
\emph{alternating transformation} for an ordered logic program $(\Pi,<)$ as
\(
\APXdefault{}=\CPX{\Pi}{<}{\CPXdefault}
\).
A fixpoint of \APOdefault{} is called an \emph{alternating fixpoint} of
$(\Pi,<)$.
Given that \CPOdefault is anti-monotonic~\cite{schwan01b}, we get that
\APXdefault{} is monotonic.
According to results tracing back to Tarski~\cite{tarski55},
this implies that \APOdefault{} possesses a least and a greatest fixpoint,
denoted by $\lfp{\APOdefault}$ and $\gfp{\APOdefault}$, respectively.

Different semantics of ordered logic programs are obtained by distinguishing
different subsets of the respective set of alternating fixpoints.
In fact, the preferred answer set semantics constitute instances of the
overall framework.
To see this, observe that each fixpoint of \CPOdefault{} is also a fixpoint of
\APOdefault.


\section{Preferring least alternating fixpoints?}
\label{sec:pwfs}

Let us now investigate the least alternating fixpoint of \APOdefault and with
it the comportment of the previous fixpoint operator in the setting of
\emph{well-founded} semantics.
As opposed to answer sets semantics,
this semantics relies on 3-valued models (or, partial models).
Such a model consists of three parts: 
the set of  true   literals,
the set of false   literals, and
the set of unknown literals.
Given that the union of these three sets is $\Lit$, it is sufficient to specify
two of the three sets for determining a 3-valued interpretation.
Accordingly, a 3-valued interpretation $I$ is a pair $(X,Y)$ where $X$ and
$Y$ are sets of literals with $X\cap Y=\emptyset$. 
That is, $L\in X$ means that $L$ is true in $I$,
while $L\in Y$ means that $L$ is false in $I$.
Otherwise, $L$ is considered to be unknown in $I$.

Well-founded semantics constitutes another major semantics for logic programs.
In contrast to answers sets semantics, it aims at characterizing skeptical
conclusions comprised in a single so-called \emph{well-founded} model of the
underlying program.
This model can be characterized within the alternating fixpoint
theory in terms of the least fixpoint of operator $A_\Pi$.
That is,
the well-founded model of a program $\Pi$ is given by the 3-valued
interpretation
\(
(\lfp{A_\Pi},\Lit\setminus C_\Pi{\lfp{A_\Pi}})
\).
Hence, it is sufficient to consider the least alternating fixpoint of a
program, since it determines its well-founded model.
We therefore refer to the least alternating fixpoint of $\Pi$ as the
\emph{well-founded set} of $\Pi$.
The set $\Lit\setminus C_\Pi{\lfp{A_\Pi}}$ is usually referred to as the
\emph{unfounded} set of $\Pi$.

After extending these concepts to preference handling,
that is,
substituting the classical operators $A_\Pi$ and $C_\Pi$ by \APOdefault{} and
\CPOdefault{}, respectively, one can show that
(i)
each ordered logic program has a unique preferred well-founded model;
(ii)
the preferred well-founded set is contained in any preferred answer set (while
the unfounded one is not);
and (iii)
whenever we obtain a two-valued well-founded model,
its underlying well-founded set is the unique answer set of the program.%
\footnote{No matter whether we consider the fixpoint operators for the
  approach in \cite{wazhli00}, \cite{breeit99a}, or \cite{descto00c},
  respectively.}

One often criticized deficiency of the standard well-founded model is that it
is too skeptical.
Unfortunately, this is \emph{not} remedied by alternating the fixpoint operators
of the previous sections, no matter which strategy we consider.
To see this, consider the ordered logic program $(\Pi_{\ref{ex:circle}},<)$:
\begin{equation}
  \label{ex:circle}
\begin{array}[t]{rcrcl}
r_1 & = & a & \leftarrow& \naf b \\
r_2 & = & b & \leftarrow& \naf a
\end{array}
  \qquad\qquad
  r_2< r_1
\end{equation}
The well-founded model of $\Pi_{\ref{ex:circle}}$ is given by
$(\emptyset,\emptyset)$.
The same model is obtained by alternating operator \CPO{\Pi_{\ref{ex:circle}}}{<}.
Observe that
\(
\CPX{\Pi_{\ref{ex:circle}}}{<}{\emptyset}=\{a, b\}
\)
and
\(
\CPX{\Pi_{\ref{ex:circle}}}{<}{\{a, b\}}=\emptyset
\).
Consequently, $\emptyset$ is the least alternating fixpoint of
$(\Pi_{\ref{ex:circle}},<)$.

The question is now why these operators are still too skeptical in defining
well-founded semantics (although they work nicely in the setting of answer
sets and regular semantics).
In fact,
the great advantage of a setting like that of answer sets semantics is that we
deal with \emph{direct} fixpoint equations, like $\CPXdefault=X$, where the
context $X$ represents the putative answer set.
This is different in the setting of well-founded semantics, where we usually
start by applying an operator to a rather small context, eg.\ initially the
empty set;
this usually results in a larger set, sometimes even \Lit, that
constitutes then the context of the second application of the operator.
Now, looking at the underlying definitions,
we see that the actual preference handling condition,
eg.\ Condition~II in Definition~\ref{def:Tp:W} takes advantage of $X$ for
deciding applicability.
The alternating character in the well-founded setting does not support this
sort of analysis since it cannot provide the (putative) final result of the
computation.


\section{Towards a preferred well-founded semantics}
\label{sec:wfs:strong}

In view of the failure of the above fixpoint operator(s) in the setting of
well-founded semantics, 
the obvious question is now whether an appropriate \emph{alternating} fixpoint
operation is definable that yields a reasonable well-founded semantics for
ordered logic programs.
As informal guidelines, we would like that the resulting semantics
(i)   allows for deriving more conclusions than the standard well-founded
      semantics by appeal to given preferences;
(ii)  coincides with standard well-founded semantics in the absence of
      preferences;
and finally
(iii) approximates the previous preferred answer sets semantics.

The standard well-founded model is defined by means of the least fixpoint of
the operator $A_\Pi=C_\Pi C_\Pi$.
As above, we aim at integrating preferences by elaborating upon the
underlying immediate consequence operator \T{\Pi,Y}{X} given
in~(\ref{eq:def:extended:immediate:consequence:operator}).
As well, the basic idea is to modify this operator so that more conclusions can
be derived by employing preferences.
However, as discussed at the end of the previous section,
the alternating iterations of $C_\Pi$ face two complementary situations:
those with smaller contexts and those with larger ones.
Since preferences exploit these contexts, it seems reasonable to distinguish
alternating applications or, at least, to concentrate on one such situation
while dealing with the other one in the standard way.%
\footnote{Such an approach is also pursued in~\cite{brewka96a}.}
For strengthening $A_\Pi=C_\Pi C_\Pi$, we thus have two options:
either
we make the outer operator derive more literals
or
we make the inner operator derive less literals.

In what follows,
we adopt the former option and elaborate upon the outer operator.
The general idea is then to reduce the context considered in the second
application of $C_\Pi$ by appeal to preferences in order to make more rules
applicable.
For this purpose we remove those literals that are derived by means of less
preferred, defeated rules.
%
\begin{definition}\label{def:alt:fixpoint:wfs:defeat}
Let $(\Pi,<)$ be an ordered logic program and let $X$ and $Y$ be sets of literals.

We define the set of 
immediate consequences of $X$ with respect to $(\Pi,<)$ and $Y$
as
\[
\TPN{\Pi}{<}{Y}{X}=
  \{
  \head{r}\mid
               r\in\Pi
                    \text{ is active wrt } (X,Y\setminus D_{(X,Y)}^r)
  \}
\]
where
\[
  D_{(X,Y)}^r
=
\left\{
L\;\left|\begin{array}{l}
                 \text{ for all rules } 
                 r'\in\Pi,\\
                 \begin{array}{l}
                   \text{ if } 
                   L=\head{r'}
                   \quad\text{ and }\\
                   \phantom{if}
                   \pbody{r'}\subseteq Y,\\
                   \text{ then }
                   r'<r
                   \quad\text{ and } \\
                   \phantom{then}
                                                 (\head{r}\cup X)\cap\nbody{r'}\neq\emptyset
                 \end{array}
               \end{array}
  \right\}\right.
\]
if $X$ is consistent, and $\TPN{\Pi}{<}{Y}{X}=\Lit$ otherwise.
\end{definition}
%
We say that $r$ defeats $r'$ wrt $X$ if
\(
(\head{r}\cup X)\cap\nbody{r'}\neq\emptyset
\).
The set of removed literals $D_{(X,Y)}^r$ consists thus of those rule heads,
all of whose corresponding rules are less preferred than $r$ and defeated by
$r$ or $X$, viz.\ the literals derived so far.
In fact, this condition only removes a literal such as \head{r'} from $Y$, if
all of its applicable generating rules like $r'$ are defeated by the
preferred rule $r$.
Note that $D_{(X,Y)}^r$ is normally different for different rules $r$.

For illustration consider the rules in~$\Pi_{\ref{ex:circle}}$.
For $X=\emptyset$ and $Y=\{a,b\}$, we get
\(
D_{(\emptyset,\{a,b\})}^{r_1}=\{b\}
\)
and
\(
D_{(\emptyset,\{a,b\})}^{r_2}=\emptyset
\).
In such a situation,
activeness of $r_1$ is checked wrt
\(
(\emptyset,\{a,b\}\setminus\{b\})
\)
while that of $r_2$ is checked wrt
\(
(\emptyset,\{a,b\})
\).
When applying $r_1$,
the removal of  $D_{(\emptyset,\{a,b\})}^{r_1}=\{b\}$ from context $\{a,b\}$ allows us
to discard the conclusion of the less preferred rule $r_2$ that is defeated by
the preferred rule $r_1$.
This example is continued below.

Notably, the choice of $D_{(X,Y)}^r$ is one among many options.
Unfortunately, it leads beyond the scope of this paper to investigate the
overall resulting spectrum, so that we concentrate on the above definition and
discuss some alternatives at the end of this section.
From a general perspective,
the above definition offers thus a parameterizable framework for
defining well-founded semantics including preferences.

In analogy to the previous sections, 
we can define a consequence operator as follows. 
%
\begin{definition}\label{def:fixpoint:operator:wfo:N}
Let $(\Pi,<)$ be an ordered logic program and let $X$ be a set of literals.

We define
\(
\CPNXdefault=\bigcup_{i\geq 0}\TPNi{i}{\Pi}{<}{X}{\emptyset}
\).
\end{definition}

Of particular interest in view of an alternating fixpoint theory is that
\CPNOdefault\ enjoys \emph{anti-monotonicity}:
%
\begin{theorem}\label{thm:CPN:anti:monotonicty}
  Let $(\Pi,<)$ be an ordered logic program
  and
  $X_1,X_2$ sets of literals.

  If $X_1\subseteq X_2$,
  then
  \(
  \CPNX{\Pi}{<}{X_2}\subseteq\CPNX{\Pi}{<}{X_1}
  \).  
\end{theorem}

Given this, 
we may define a new alternating transformation of $(\Pi,<)$ as
\[
\APNOdefault{}=\CPNOdefault{}C_\Pi.
\]
Since both $\CPNOdefault{}$ and $C_\Pi$ are anti-monotonic,
$\APNOdefault{}$ is monotonic.
%
\begin{definition}\label{def:alt:fixpoint:wfs:set:N}
Let $(\Pi,<)$ be an ordered logic program and let $X$ be a set of literals.

We define $X$ as a preferred well-founded set of $(\Pi,<)$ iff
\(
\lfp{\APNOdefault}=X
\).
\end{definition}

By Tarski's Theorem~\cite{tarski55}, we get that each ordered logic
program has a unique preferred well-founded set.
%
\begin{theorem}\label{thm:alt:fixpoint:unique}
Let $(\Pi,<)$ be an ordered logic program. 

Then, there is a unique preferred well-founded set of $(\Pi,<)$.
\end{theorem}
%
Given the notion of the preferred well-founded set,
we define the preferred well-founded model of an ordered program as follows.
%
\begin{definition}\label{def:alt:fixpoint:wfs:W}
  Let $(\Pi,<)$ be an ordered logic program
  and  
  let $X$ be the well-founded set of $(\Pi,<)$.
  
  We define the preferred well-founded model of $(\Pi,<)$
  as
  \(
  (X,\Lit\setminus C_\Pi(X))
  \).
\end{definition}
%

It is well-known that the standard well-founded semantics for extended
logic programs has time complexity $O(n^2)$~\cite{wit91,bargel94a}.
The complexity of the preferred well-founded semantics is still in 
polynomial time but it is in $O(n^3)$.
The reason is that we have to additionally compute $D_{(X,Y)}^r$
for each $r\in\Pi$.

%

We first obtain the following corollary to Theorem~\ref{thm:alt:fixpoint:unique}.
%
\begin{corollary} \label{thm:alt:wfm:unique}
  Every ordered logic program has a unique preferred well-founded model.
\end{corollary}
%
This result shows that our preferred well-founded semantics is as robust as
the standard well-founded semantics.

The relationship between the standard well-founded model and the preferred 
well-founded model can be stated as follows.
%
\begin{theorem}\label{thm:relationship:wfm}
Let $(X,Y)$ be the preferred well-founded model of $(\Pi,<)$
and 
let $(X',Y')$ be the well-founded model of $\Pi$. 

Then, we have
\begin{enumerate}
\item $X'\subseteq X$ and $Y'\subseteq Y$ and
\item $(X,Y)=(X',Y')$, if ${<}=\emptyset$.
\end{enumerate}
\end{theorem}

Let us reconsider $(\Pi_{\ref{ex:circle}},<)$.
While $(\emptyset,\emptyset)$ is the well-founded model of $\Pi_{\ref{ex:circle}}$,
its ordered counterpart $(\Pi_{\ref{ex:circle}},<)$ has the preferred
well-founded model $(\{a\},\{b\})$.
To see this, observe that
\(
C_{\Pi_{\ref{ex:circle}}}{\emptyset}=\{a,b\}
\)
and
\(
\CPNX{\Pi_{\ref{ex:circle}}}{<}{\{a,b\}}=\{a\}
\).
Clearly, $\{a\}$ is a fixpoint of $C_{\Pi_{\ref{ex:circle}}}$ and
$\CPNO{\Pi_{\ref{ex:circle}}}{<}$.
Thus, $\{a\}$ is an alternating fixpoint of $(\Pi_{\ref{ex:circle}},<)$.
Also, we see that $\emptyset$ is not an alternating fixpoint.
This implies that $\{a\}$ is the least alternating fixpoint of
$(\Pi_{\ref{ex:circle}},<)$.

This example along with the last result show that preferences allow us to
strengthen the conclusions obtained by the standard well-founded semantics.
That is, whenever certain conclusions are not sanctioned in the standard
framework one may add appropriate preferences in order to obtain these
conclusions within the overall framework of well-founded semantics.

For a complement, consider the following variation of
$(\Pi_{\ref{ex:circle}},<)$, also discussed in~\cite{brewka96a}.
\begin{equation}
  \label{ex:wfs:benchmark}
  \begin{array}[t]{rcrcl}
    r_1 & = & a & \leftarrow& \naf b \\
    r_2 & = & b & \leftarrow& \naf c
  \end{array}
  \qquad\qquad
  r_2< r_1
\end{equation}
Observe that $\Pi_{\ref{ex:wfs:benchmark}}$ has well-founded model $(\{b\},\{a,c\})$.
In contrast to  $(\Pi_{\ref{ex:circle}},<)$,
the preferred well-founded model of $(\Pi_{\ref{ex:wfs:benchmark}},<)$ is also
$(\{b\},\{a, c\})$.
As discussed in~\cite{brewka96a} this makes sense since preferences should
only enrich but not ``override'' an underlying well-founded model.

Another attractive property of this instance of preferred well-founded
semantics is that it provides an approximation of preferred answer sets
semantics.
%
\begin{theorem}\label{thm:pwfs:approx:B}
Let $(X,Y)$ be the preferred well-founded model of $(\Pi,<)$
and 
let $Z$ be a preferred answer set of $(\Pi,<)$. 

Then, we have
$X\subseteq Z$ and $Y\subseteq \Lit\setminus Z$.
\end{theorem}
%
Notably,
this can be shown for all aforementioned preferred answer sets semantics,
no matter whether we consider the approach in \cite{wazhli00},
\cite{breeit99a}, or \cite{descto00c}, respectively.

Finally, let us briefly discuss some alternative choices for $D_{(X,Y)}^r$.  In
fact, whenever we express the same preferences among (negative)
rules having the same
head the previous definition of $D_{(X,Y)}^r$ is equivalent to
\(
\{\head{r'}\mid r'<r\text{ and }(\head{r}\cup X)\cap\nbody{r'}\neq\emptyset\}
\).
However, this conceptually simpler definition is inadequate when it comes to
attributing different preferences to rules with the same heads as in the
following example.

Consider the ordered program $(\Pi_{\ref{ex:counterexa:aug}},<)$.
\begin{equation}
  \label{ex:counterexa:aug}
  \begin{array}[t]{rcrcl}
    r_1 & = & a & \leftarrow& \\
    r_2 & = & b & \leftarrow& \naf a \\
    r_3 & = & a & \leftarrow& \naf b
  \end{array}
  \qquad\qquad
  r_3<r_2< r_1
\end{equation}
The preferred well-founded semantics of $(\Pi_{\ref{ex:counterexa:aug}},<)$
gives $(\{a\},\{b\})$,
while the conceptually simpler one yields $(\{a,b\},\emptyset)$,
a clearly wrong result!
In the simplistic setting $D_{(\emptyset,\{a,b\})}^{r_2}$ would contain the head of the
third rule, discarding the fact that $r_1$ already defeats $r_2$.

Another alternative choice for $D_{(X,Y)}^r$ is indicated by the difference between
the strategies employed in \cite{wazhli00} and \cite{descto00c}.
In fact, the latter implicitly distinguishes between same literals stemming
from different rules.
This amounts to distinguishing different occurrences of literals.
For this, we may rely on the aforementioned simplistic definition of $D_{(X,Y)}^r$
and suppose that \head{r} provides us with occurrences of literals, like
$b^{r_2}$ instead of $b$.
Without entering details, let us illustrate this idea by appeal to
$(\Pi_{\ref{ex:counterexa:aug}},<)$.
An approach distinguishing occurrences of literals would yield
\(
C_{\Pi_{\ref{ex:counterexa:aug}}}{\emptyset}=\{a^{r_1},b^{r_2},a^{r_3}\}
\)
and
\(
\CPNX{\Pi_{\ref{ex:counterexa:aug}}}{<}{\{a^{r_1},b^{r_2},a^{r_3}\}}=\{a^{r_1},a^{r_3}\}
\).
When considering $r_2$, we check activeness wrt
\(
(\emptyset,\{a^{r_1},b^{r_2},a^{r_3}\}\setminus\{a^{r_3}\})
\),
viz.\
\(
(\emptyset,\{a^{r_1},b^{r_2}\})
\).
Unlike just above,
$a^{r_1}$ remains in the reduced context and $r_2$ is inapplicable.
An elaboration of this avenue is beyond the scope of this paper, in
particular, since it involves an occurrence-based development of well-founded
semantics.


\section{Relationships}
\label{sec:relationships}

In contrast to answer set semantics,
the extension of well-founded semantics to ordered logic program has been
rarely studied before.
In this section we will discuss the relation of our approach to
\cite{brewka96a,nute87,yowayu01}.

\subsection{Relation to Brewka's Approach}

Brewka defines in~\cite{brewka96a} a well-founded semantics for ordered logic
programs.
Notably, this approach is based on a paraconsistent extension of well-founded
semantics that tolerates inconsistencies among the result of the inner operator
without trivializing the overall result.
Despite this deviation from standard well-founded semantics,
the question remains whether Brewka's semantics can be captured within our
semantical framework.
%

In fact, both approaches are based on quite different intuitions.
While the underlying idea of Brewka's approach is to define a criterion for
selecting the intended rules by employing preference,
we integrate preferences into the immediate consequence operaor by
individually restricting the context of application for each rule.

Nonetheless, it turns out that Brewka's semantics can be captured through an
alternating fixpoint construction.
As we show below, Brewka's modification boils down to using an alternate
fixpoint operator of the form
``\(
{\mathcal{C}^\star_{(\Pi,<)}}{{C}^\star_{\Pi}}
\)''.
To this end, let us first consider the difference among the underlying
operators ${{C}^\star_{\Pi}}$ and $C_\Pi$.
Define  $\Cl{\Pi}$ as the smallest set of literals which is closed under a
basic program $\Pi$.
Then, given a set $X$ of literals, ${{C}^\star_{\Pi}(X)}$ is defined as $\Cl{\Pi^X}$.
Dropping the requirement of logical closure results in a paraconsistent
inference operation.
For example, given
\(
\Pi=\{a\LPif,\neg a\LPif,b\LPif\}
\),
we get
$\Cn{\Pi}=\Lit$, while $\Cl{\Pi}=\{a,\neg a,b\}$.
Although the corresponding adaptions are more involved, the surprising result
is now that Brewka's semantics can also be captured within our overall
framework, if we use the closure operator $\ClO$ instead of $\CnO$.

Moreover, we need the following.
%
Let $(\Pi,<)$ be an ordered logic program and $X$ be a set of literals.
We define $\Pi_X^r$ as  
the set of rules defeated by $r$ wrt $X$ and $<$ as
\[
\Pi_X^r
=
\{r'\in\Pi\mid r'<r, \; r \text{ defeats } r' \text{ wrt } X\}.
\]
%
Notice that $\Pi_X^r$ is a set of rules while $D_{(X,Y)}^r$ is a set of literals.
$\Pi_X^r$ is also different from Brewka's \emph{Dom} (set of \emph{dom}inated rules)
in that $\Pi_X^r$ is defined wrt a set $X$ of literals rather
than a set of rules.

Write $(\Pi_X^r)^+=\{(r')^+\mid r'\in \Pi_X^r\}$.
Let ${\mathcal{T}^\star_{(\Pi,<)}}$ be the operator obtained from
${\mathcal{T}^\circ_{(\Pi,<)}}$ (in Definition~\ref{def:alt:fixpoint:wfs:defeat})
by replacing $Y\setminus D_{(X,Y)}^r$ with $\Cl{{\Pi^Y\setminus (\Pi_X^r)^+}}$.
This results in a fixpoint operator ${\mathcal{C}^\star_{(\Pi,<)}}$.

As we show in the full version of this paper,
Brewka's well-founded set corresponds to the least fixpoint of the alternating operator 
\(
{\mathcal{C}^\star_{(\Pi,<)}}{{C}^\star_{\Pi}}
\).
This means Brewka's well-founded semantics also enjoys an alternating fixpoint
characterization.
%
%
%

\subsection{Relation to Other Approaches}
In \cite{yowayu01}, it is mentioned that a well-founded semantics with
preference can be defined in terms of their operator but default negation is
not allowed in their syntax.
However, even for ordered logic programs without default negation, our basic
semantic approach is different from the well-founded semantics in priority
logic~\cite{yowayu01}.
The main reason is that they interpret the priority relation $r< r'$ in a
quite different way: $r$ is blocked whenever $r'$ is applicable.
While we attribute to the program
\begin{equation}
  \label{ex:you}
  \begin{array}[t]{rcrcl}
    r_1 & = & p & \leftarrow &\\
    r_2 & = & q & \leftarrow &
  \end{array}
  \qquad\qquad
  r_2< r_3
\end{equation}
a preferred well-founded model, containing both $p$ and $q$,
the well-founded model of $\Pi_{\ref{ex:you}}$ in priority logic is $\{p\}$.
That is, $q$ cannot be inferred. 

Another skeptical semantics for preference is defeasible logic,
which was originally introduced by D. Nute~\cite{nute87} and received
extensive studies in recent years~\cite{anbigo00,anbigo01,maroan00}.
Defeasible logic distinguishes the strict rules from defeasible rules.
This already makes its semantics different from our preferred well-founded
semantics. 

Consider an example from~\cite{brewka01}. The following is a theory in
defeasible logic:

\begin{equation}
  \label{ex:brewka5}
  \begin{array}[t]{rcrcl}
    r'_1 & &\Rightarrow & p\\
    r'_2 &p&\rightarrow & q\\
    r'_3 & &\Rightarrow & \neg q 
  \end{array}
  \qquad\qquad
  r'_2< r'_3
\end{equation}
In defeasible logic, $+\delta q$ is not derivable, i.~e., $q$ cannot
be defeasibly derived. As pointed out by Brewka, this means a defeasible
rule having higher priority can defeat a strict rule. 

The above theory can be directly translated into an ordered logic program
$(\Pi,<)$ as follows:

\begin{equation}
  \begin{array}[t]{rcrcl}
    r_1 & = & p      & \leftarrow & \naf\neg p\\
    r_2 & = & q      & \leftarrow & p         \\
    r_3 & = & \neg q & \leftarrow & \naf q
  \end{array}
  \qquad\qquad
  r_2< r_3
\end{equation}
It can be verified that the preferred well-founded model (in our sense)
is $\{p,q\}$. Therefore, $q$ is derivable under our preferred well-founded
semantics.

\section{Conclusion}
\label{sec:conclusion}

We have looked into the issue of how van Gelder's alternating fixpoint 
theory~\cite{vangelder93} for normal logic programs can be suitably
extended to define the well-founded semantics for ordered logic programs
(extended logic programs
with preference). 
The key of the alternating fixpoint approach is how to specify
a suitable consequence relation for ordered logic programs.
We argue that the preference strategies for defining
answer sets are not suitable for defining preferred well-founded
semantics and then some informal criteria for
preferred well-founded semantics are proposed.
Based on this analysis, we have defined a well-founded semantics for ordered
logic programs. This semantics allows an elegant definition and satisfies
some attractive properties: (1) Each ordered logic program has a unique
preferred well-founded model; (2) The preferred well-founded reasoning
is no less skeptical than the standard well-founded reasoning;
(3) Any conclusion under the preferred well-founded semantics is also
derivable under some major preferred answer sets semantics. 
Our semantics is different from defeasible
logic and the skeptical priority logic.
An important result is the equivalence of Brewka's preferred well-founded
semantics and our semantics introduced in Section~\ref{sec:wfs:strong}.

\paragraph{Acknowledgments.}
This work was supported by DFG under grant FOR~375/1-1,~TP~C.


\end{document}